\documentclass[sigconf]{acmart}
\usepackage{subfig}
\AtBeginDocument{%
  \providecommand\BibTeX{{%
    \normalfont B\kern-0.5em{\scshape i\kern-0.25em b}\kern-0.8em\TeX}}}

\copyrightyear{2021}
\acmYear{2021}
\setcopyright{acmcopyright}

\acmConference[MMAsia '21]{ACM Multimedia Asia}{December 1--3, 2021}{Gold Coast, Australia}
\acmBooktitle{ACM Multimedia Asia (MMAsia '21), December 1--3, 2021, Gold Coast, Australia}
\acmPrice{15.00}
\acmDOI{10.1145/3469877.3490611}
\acmISBN{978-1-4503-8607-4/21/12}

\acmSubmissionID{46}

\begin{document}

\title{Patch-Based Deep Autoencoder for Point Cloud Geometry Compression}


\author{Kang You}
\affiliation{%
  \institution{Nanjing University of Aeronautics and Astronautics}
  \city{Nanjing}
  \country{China}
}
\email{youkang@nuaa.edu.cn}

\author{Pan Gao}
\affiliation{%
  \institution{Nanjing University of Aeronautics and Astronautics}
  \city{Nanjing}
  \country{China}}
\email{pan.gao@nuaa.edu.cn}
\authornote{Corresponding author}


\begin{abstract}
    The ever-increasing 3D application makes the point cloud compression unprecedentedly important and needed. In this paper, we propose a patch-based compression process using deep learning, focusing on the lossy point cloud geometry compression. Unlike existing point cloud compression networks, which apply feature extraction and reconstruction on the entire point cloud, we divide the point cloud into patches and compress each patch independently. In the decoding process, we finally assemble the decompressed patches into a complete point cloud. In addition, we train our network by a patch-to-patch criterion, i.e., use the local reconstruction loss for optimization, to approximate the global reconstruction optimality. Our method outperforms the state-of-the-art in terms of rate-distortion performance, especially at low bitrates. Moreover, the compression process we proposed can guarantee to generate the same number of points as the input. The network model of this method can be easily applied to other point cloud reconstruction problems, such as upsampling.
\end{abstract}




\maketitle

\section{Introduction}
	\label{sec:Introduction}
	3D point cloud is an essential data structure in 3D representation, which has received increasing attention due to the popularity rise of Virtual Reality and Mixed Reality \cite{MRapp}. A point cloud is a set of points in the three-dimensional space, and each point is specified by $(x,y,z)$ coordinates and optional attributes. The amount of point cloud data is usually large, which places high requirements on the point cloud compression (PCC) methods.
	
	The research on point cloud compression can be roughly divided into three main categories \cite{MPEGStandards}: LIDAR point cloud compression (L-PCC), surface point cloud compression (S-PCC), and video-based point cloud compression (V-PCC). In this paper, we focus on the lossy geometry encoding and decoding process of the S-PCC. However, traditional lossy PCC methods without using deep learning generally struggle with the performance at low bitrates. For example, the number of points generated by octree based compression methods \cite{Octree, Octree2} will decrease abruptly as the tree depth is lowered. It also generates blocky results similar to the mosaic effect.
	
	Autoencoder is a new machine learning based model to deal with data compression \cite{VariationalCompression, EndToEnd}, which can automatically learn the analysis and synthesis transforms tailed for point cloud data. However, most of the existing autoencoder methods of point cloud compression are based on voxelization and 3D convolution \cite{Quach,Quach2,JWang}, which causes inefficiency in the usage of memory and time. Besides, they cannot perform effectively on irregular and sparse point clouds. Recently, PointNet \cite{PointNet} and PointNet++ \cite{PointNet++} are proposed to extract point cloud features directly from original points without voxelization. Following this trend, some works  \cite{LRepresentations,DAE} consider to apply PointNet and multilayer perceptron (MLP) to compress point clouds. These methods perform well on some sparse shapes, but they are almost impossible to compress dense point clouds due to the high-dimensional fully connected output, and they will induce the missing of the structure and details for complex shapes.
	
	In this paper, we propose an end-to-end autoencoder based on patches. Inspired from PointNet for classification and segmentation, we design a new network model for point cloud compression. The proposed model comprises an analysis transform for transforming the point cloud data to a global feature, a uniform quantizer to quantize the latent feature, and a synthesis transform to transform back to the data space. Instead of taking a whole complex point cloud as the input of neural network directly, we divide the point cloud into several patches and compress each patch independently. In the decoding process, each patch is reconstructed separately, and then the patches are combined by using coordinate information of the sampling points.
	
	The idea of using patch for training the model has two merits. Firstly, as the patch is already a local region of the point clouds, the network model does not need to be very deep or using multiple stacks of set abstraction layers to capture fine details, which thus reduces the training model complexity. As will be demonstrated, our proposed model is easy to optimize and train while enjoying coding performance gains. Secondly, dividing the point cloud into patches augments the training data, which can avoid the overfitting problem and improve the model prediction accuracy. We trained our network on the ModelNet40 dataset \cite{ModelNet40}, and test its performance on ModelNet40 and ShapeNet \cite{ShapeNet}. Then, we compare it with Quach’s method \cite{Quach}, Yan's method \cite{DAE}, Octree \cite{Octree}, and TMC13 \cite{TMC13}. We find that our method can compress the geometry data of point clouds efficiently while having minimum quality loss. Additionally, our method has significantly lower GPU memory usage compared to voxelization-based networks.
	
	In the next section, we will review some related works. After describing the proposed method in Section \ref{sec:ProposedMethod}, we show our experiment results in Section \ref{sec:ExperimentalResults}. Conclusions are drawn in Section \ref{sec:Conclusion}.

\section{Related Work}
\label{sec:Relatedwork}
	Generally speaking, there are mainly three kinds of methods for point cloud geometry compression task: traditional compression algorithm without deep learning, voxel-based autoencoder, and PointNet based autoencoder.
	
	The octree based compression \cite{Octree} is the most representative method of traditional point cloud geometry compression. The octree algorithm recursively subdivides the coordinate space of the point cloud to generate an octree structure. At the lowest level, each subspace is used as a leaf node of the octree, and then the tree is encoded. TMC13 \cite{TMC13} is a compression platform recently proposed by the MPEG organization, which contains octree based and other nested-partition based compression algorithms.
	
	Quach \emph{et al.} proposed an autoencoder using voxels as input and 3D convolution as the backbone in \cite{Quach}. Their first step is voxelization, that is, the $N\times3$ matrix is transformed into a three-dimensional binary matrix. In the next step, the point cloud is done with 3D convolution multiple times to extract the hidden layer feature, and then the hidden layer feature is quantized and entropy coded. Although this method has achieved excellent results, it is limited to the compression of voxelized point clouds. In addition, most computations in 3D convolution are redundant because many 3D voxel space are unoccupied.
	
	PointNet is a kind of neural network structure that directly takes the point cloud coordinate set as the input. The authors of \cite{PointNet} propose a symmetric function to extract features from unordered point sets. The symmetric function specifically is implemented using the shared-MLP and max pooling. Finally, the segmentation and classification results are calculated through further operation in the feature domain. However, PointNet lacks the process to obtain local features, and, to overcome this issue, PointNet++ \cite{PointNet++} was born. Pointnet++ uses several repetitive operations of sampling, grouping, and PointNet to extract features from the neighborhood at different scales of a point cloud, which is similar to hierarchical convolution operation in 2D images. It has achieved remarkable results in classification and segmentation tasks. Inspired by the success of those point-based models, the authors of \cite{DAE} used an autoencoder based on PointNet to compress point cloud data, which takes as input the points rather than voxel grids. In this point-based autoencoder, the PointNet is used directly as the encoder to compress the whole point cloud, and the fully connected layer is employed as the decoder. However, this naive implementation of the compression model using PointNet will result in exponentially growing computational cost when the point counts become large. 
	
	The main differences of this paper with the above work can be summarized as:
	\begin{itemize}
		\item We tried a method of dividing a point cloud into patches. Different from two-dimensional image, the points of a point cloud are independent of each other, and there is no regular voxel-space correlation or connectivity. For better local structure capture, we carefully calculate the total patch count and the number of points in a patch, and generate the patches using point sampling and KNN operation.
		\item We proposed a patch-based neural network architecture to compress point clouds. It can yield a high compression ratio with acceptable loss in reconstruction. It can also freely adjust the number of points of the reconstructed point cloud, which benefits greatly to other point-based reconstruction problem, such as upsampling.
	\end{itemize}

	\section{Proposed Method}
	\label{sec:ProposedMethod}
	
	\begin{figure*}
		\includegraphics[width=\textwidth]{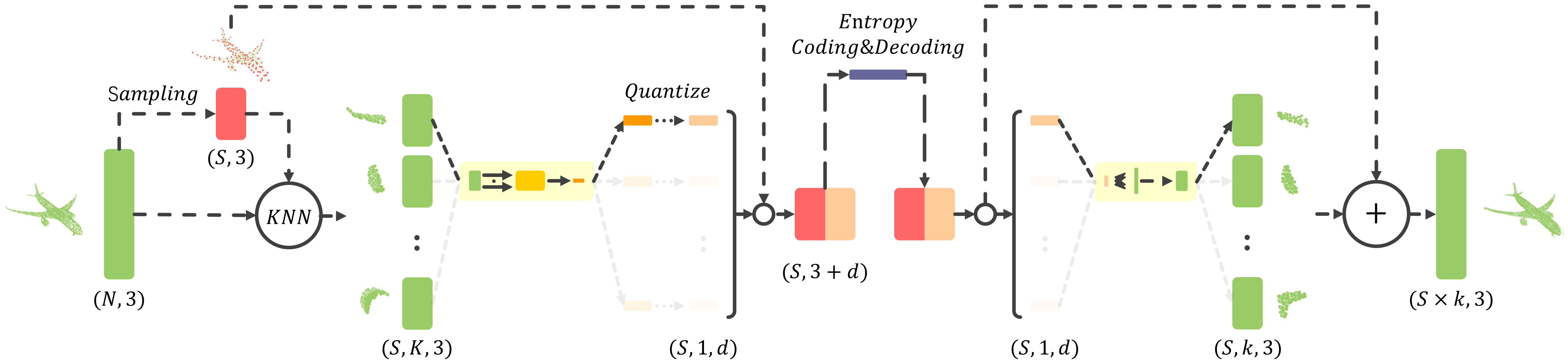}
		\caption{Compression process. The point cloud is divided into $S$ patches with $K$ points for each. In the encoding process, the patch features and sampled points are concatenated to a final latent representation of size $\left ( S, 3+d \right )$. In the decoding process, the final representation are separated into the same two parts as the encoder. Decoded patches are finally assembled to produce a point cloud of size $(S \times k, 3)$ using the coordinates of sampled points.}
		\label{fig:comp_process}
	\end{figure*}
	
	\begin{figure*}
		\includegraphics[width=\textwidth]{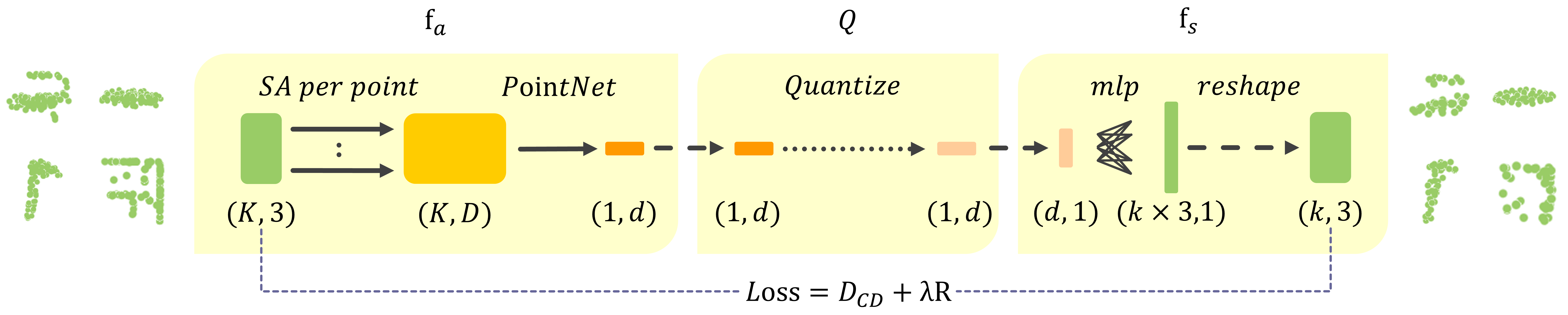}
		\caption{The proposed autoencoder architecture. In this figure, solid arrows represent neural network operations with back-propagation while dashed arrows represent arithmetic operations without back-propagation, and dotted arrows particularly represent the quantization operation. Input and output are both patches, and Chamfer Distance (CD) is used to evaluate the distortion between the input and output. “mlp” means multi-layer perceptron.}
		\label{fig:nn_archi}
	\end{figure*}
	
	Figure \ref{fig:comp_process} shows the point cloud compression process using our trained autoencoder. Figure \ref{fig:nn_archi} shows the details of the proposed autoencoder for training.
	
	\subsection{Proposed Compression Process}
	In our compression process, we first divide a point cloud into two parts: independent simple patches, and the auxiliary information of the coordinate of the sampling points between patches. During the encoding process, we pass each patch into the encoder of the autoencoder, to generate a set of hidden layer representations. These representations are quantized using a uniform quantizer, which are then combined with the coordinate information of the sampling centroid points to form the final latent representation of the whole point cloud. Finally, the final latent representation is entropy encoded to a bitstream and transmitted to the decoder. The decoding process is basically the reverse process of the encoding process. After obtaining the representation of the point cloud, we separate latent patch representations from it and pass these representations to the decoder of the autoencoder separately. And then, we combine the decoder outputs with the auxiliary coordinate information to form the final reconstruction result.
	
\begin{figure}[b!]
		\centering
		\subfloat[Original]
		{\label{fig:s_r_1}\includegraphics[width=0.25\linewidth]{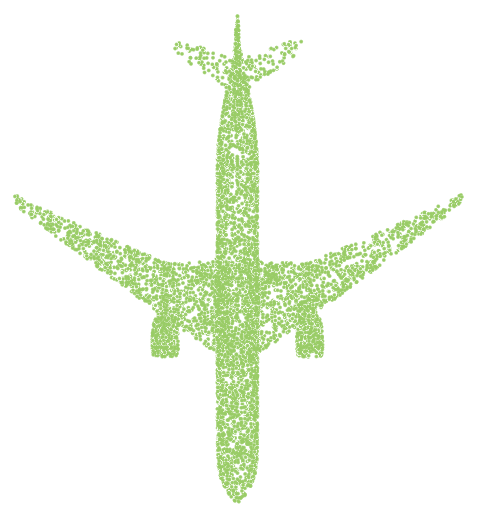}}
		\subfloat[Sampled]
		{\label{fig:s_r_2}\includegraphics[width=0.25\linewidth]{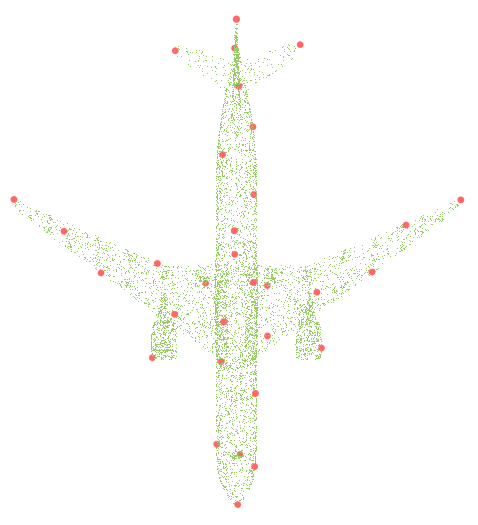}}
		\subfloat[$\alpha=1$]
		{\label{fig:s_r_3}\includegraphics[width=0.25\linewidth]{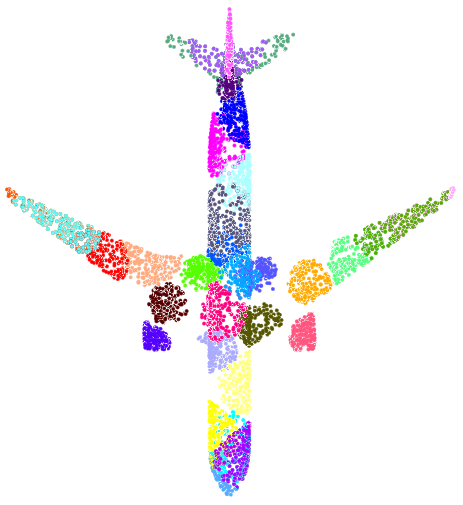}}
		\subfloat[$\alpha=2$]
		{\label{fig:s_r_4}\includegraphics[width=0.25\linewidth]{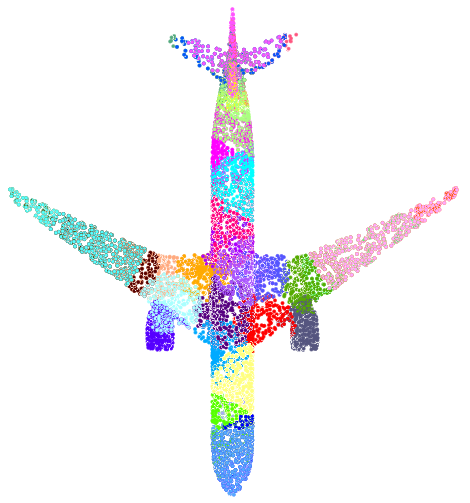}}
		\caption{An example of patch division. (a) is an 8192-point resolution point cloud. (b) is the farthest point sampling results when $S=32$. (c) represents the divided patches in the setting of $S=32,K=256$, while (d) is in $S=32,K=512$. It can be seen that point cloud structure information cannot be fully captured even when $S$ times $K$ is exactly the number of points of the input point cloud.}
		\label{fig:samp_results}
	\end{figure}

	Different from two-dimensional image segmentation, we use sampling and K nearest neighbor (KNN) to segment the point cloud into patches of the same resolution. The process is described as follows:
	
	\begin{itemize}
		\item For the original point cloud $x$ with $N$ points, we use the farthest point sampling (FPS) to sample $S$ centroid points $\left \{p_{1},p_{2},...,p_{S} \right\}$ from the original point cloud.
		\item For each sampled point $p_{i}$, we use KNN to find $K$ neighboring points $\left \{ p_{i}^{1}, p_{i}^{2}, ..., p_{i}^{K} \right \}$. By subtracting the sampled point $p_{i}$ from the $K$ points, we obtain a set of coordinate difference, i.e., $\left \{ p_{i}^{1}-p_{i}, p_{i}^{2}-p_{i}, ..., p_{i}^{K}-p_{i} \right \}$, which forms the patch to be inputted to the network model.
		\item Now we can get $S$ patches, and each patch has $K$ points. Meanwhile, the FPS results, the coordinate $\left \{p_{1},p_{2},...,p_{S} \right\}$ are saved as the auxiliary information of the patches. The coordinates of the sampling points serve as the skeleton centre points to restore the detailed points, which can thus make the output point cloud matches well with the ground truth point cloud. To determine the appropriate number of neighboring points in a patch, we conduct a series of trails. As illustrated in Figure \ref{fig:samp_results}, even making the total number of points of all patches equal to that of the input point clouds is insufficient to cover the whole point cloud. Thus, we here use $S\times K= \alpha N\left ( \alpha > 1 \right )$ to avoid the situation that some points cannot be captured. In the following, we will also investigate the effect of patch size and the patch counts on the overall coding performance.
	\end{itemize}
	
	In the decoding process, we decode the representation of each patch separately and then add it back to the sampled point $p_{i}$. In this way, we can get the prediction result. The union of each independent prediction result is the final point cloud reconstruction result. It should be noted that in the compression process, we divided the patch by satisfying $S\times K= \alpha N$, and in order to get the same resolution as the input point cloud, we set $k=K/\alpha$, where $k$ is the number of points of the network predicted point cloud patch.
	
	\subsection{Autoencoder Architecture for Training}
	We designed an autoencoder based on PointNet to implement transformation and compression for patches. It includes an analysis transform $f_{a}$, a quantization function $Q$, and a synthesis transform $f_{s}$. Analysis transform is used to extract the hidden features from a simple patch, quantization is used to quantize the hidden feature for further compression, and synthesis transform is used to reconstruct the quantized feature to the input shape.
	
	In the analysis transformation, we first use a set abstraction (SA) layer to extract a local feature at a small scale for each point. Then we use a PointNet to extract a higher level global feature. SA is developed in PointNet++, which originally consists of three steps: sampling layer, grouping layer, and PointNet layer. The ``SA per point" we use here only includes grouping points and extracting features using PointNet. After using ``SA per point", as illustrated in Fig. \ref{fig:nn_archi}, we get a point cloud patch feature matrix with each point having $D$ dimensional feature. Following this process,  a PointNet layer is inserted to extract the global feature vector $(1,d)$.
	
	As for our quantization process, inspired by \cite{EndToEnd}, we add an element-wise uniform noise between -0.5 and 0.5 to the $\left ( 1,d \right )$ dimensional hidden feature in training. Here, we use uniform noise approximation, which can make the quantization process differentiable, and thus allows for back propagation during stochastic gradient descent optimization. In testing, we use the rounding operation for the hidden layer features in order to implement subsequent entropy coding.
	
	Our synthesis transform $f_{s}$ is composed of several fully connected layers. Fully connected layer has been used in many reconstruction tasks and reached excellent results \cite{LRepresentations, DenseReconstruction}. The last step is reshaping the output of the multi-layer perceptron into point geometry matrix $k\times 3$, which is our reconstruction result of one patch.
	
	We use the Chamfer distance to constrain the error between the reconstructed results and the input patch, which is shown below:
	
	\begin{equation}
	D_{CD}=	\frac{1}{P}\sum_{i=1}^{P}D'_{CD}\left ( S_{i},S'_{i} \right )
	\end{equation}
where:	
	\begin{equation*}
		\begin{split}
		D'_{CD}\left ( S_{i},S'_{i} \right )= & \frac{1}{\left |S_{i}\right |}\sum_{x\epsilon S_{i}}\min_{y\epsilon S'_{i}}\left \| x-y \right \|_{2}^{2} \\ 
		& + \frac{1}{\left |S'_{i}\right |}\sum_{y\epsilon S'_{i}}\min_{x\epsilon S_{i}}\left \| y-x \right \|_{2}^{2}
		\end{split}
		\label{eq:cd}
	\end{equation*}
	where $P$ is the number of patches in a batch during training, $x$ represents a point from the patch $S_{i}$ from the original point cloud, and $y$ is a point from the network predicted patch $S'_{i}$. Chamfer distance can effectively measure the distance between two point sets, and it is more computationally efficient than the earth mover's distance \cite{fan2017point}.
	
	With the distortion between two point sets defined, our final loss function is set to $L= D_{CD} + \lambda R$, where $R$ is the bit rate estimated by the probability distribution of hidden layer features. The bit rate estimation expression is given as follows:
	\begin{equation}
		R=\frac{1}{P}\sum_{i=1}^{P}\left( -q(\tilde{z_i}|S_i) \cdot log_2{p_{\tilde{z_i}}(\tilde{z_i})}\right)
	\end{equation}
	where $\tilde{z_i}$ is the hidden representation after adding uniform noise for patch $S_i$. $q(\tilde{z_i}|S_i)$ denotes the actual marginal distribution of $\tilde{z_i}$, i.e., the so-called variational posterior probability in the general variational autoencoder \cite{minnen2018joint}. $p_{\tilde{z_i}}(\tilde{z_i})$ is the entropy model of $\tilde{z_i}$, which can be estimated by using a non-parametric, fully factorized density model, similar to the modeling process in \cite{EndToEnd}. What needs to be explained is the effect of $\lambda$ on compression ratio was not significant, and we adjust compression ratio mainly relying on changing the size of bottleneck $d$.
	
	Specifically, we implement our network using the following network parameters:
	
	\begin{math}
	SA\left ( K,8,\left [ 32, 64, 128 \right ] \right )\rightarrow PN\left ( \left [ 64, 32, d \right ] \right )\rightarrow Quantization\rightarrow \\ FC\left ( d,128 \right )\rightarrow ReLU\rightarrow FC\left ( 128, 256 \right )\rightarrow ReLU\rightarrow FC\left ( 256,k\times 3 \right )
	\end{math}
	
	Layers are specified using the following format:
	
	Set Abstraction: $SA$ (the number of points in a patch, the number of points in each group, shared MLP layer sizes)
	
	PointNet: $PN$ (shared MLP layer sizes)
	
	Fully Connected Layer: $FC$ (input feature size, output feature size)
	
	\section{Experimental Results}
	\label{sec:ExperimentalResults}
	We use the ModelNet40 training set (a total of 9835 shapes) to train our proposed autoencoder network. Then we perform tests on the ModelNet40 testing set (a total of 2467 shapes) and ShapeNet testing set (a total of 2874 shapes). It is important to note that the data in ModelNet40 is much more complex than ShapeNet, regardless of the size, orientation and position of the point clouds. All point cloud data is created by sampling points uniformly on each shape and their coordinates are zoomed to $\left[0, 64\right]$ for fair comparison with other related methods (e.g., Quach's \cite{Quach}).
	
	For data preparation, we use the point cloud data created by sampling 8192 points from ModelNet40 for regular training and testing, and the data by sampling 2048 points from ShapeNet to further test the robustness of our method.
	
	Particularly, as we divide each point cloud into $S$ patches, we can obtain a total of $9835 \times S$ patches for training and $2467 \times S$ patches for testing in ModelNet40. This amount of data can significantly avoid the over-fitting problem commonly encountered in model training.
	
	We implement our network on Python 3.6 and Pytorch 1.2. We use the Adam optimizer \cite{Adam} with an initial learning rate of 0.0005 and a batch size of 16. As the amount of training data in one epoch depends on the $S$, we directly set the maximum number of steps to about 40,000 during each training. For the patch division, we set $\alpha=2$, i.e., $S \times K=2N$, to cover the whole point cloud as much as possible. Finally, we use the point-to-plane symmetric PSNR \cite{p2planePSNR} to compute the reconstruction distortion.
	
	\subsection{Visualization of Training Process}
	
\begin{figure}
		\centering
		\includegraphics[width=0.375\textwidth]{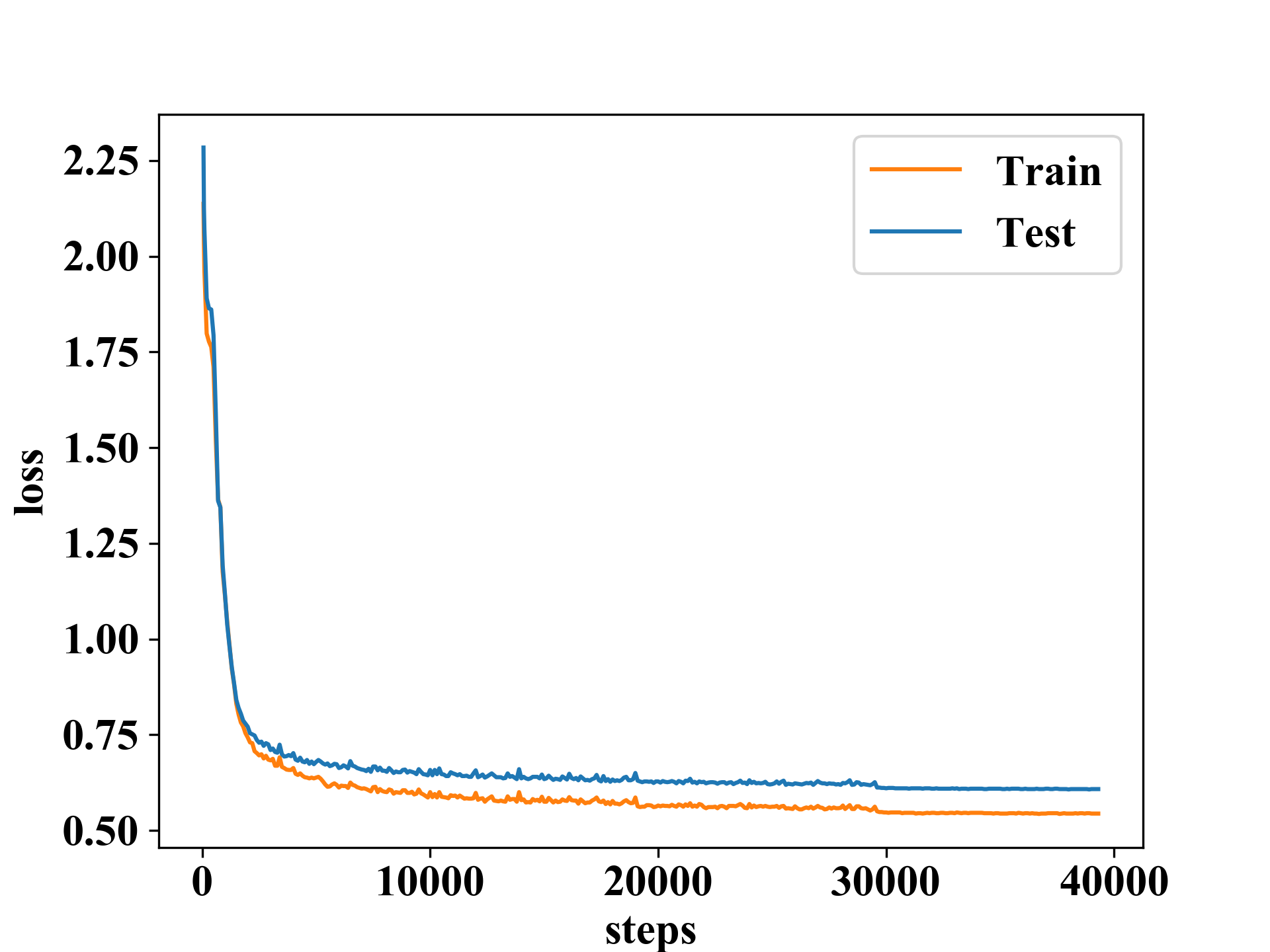}
		\caption{Loss curves during training.}
		\label{fig:Loss}
	\end{figure}

	In this subsection, we choose a few point clouds to demonstrate the training process of our network. We save the network parameters after certain steps, and then use the model with these parameters to compress selected point clouds. From Table \ref{tab:train_process}, we can see with the increase of iteration steps, the reconstructed point cloud is gradually close to the ground truth. For a better illustration of the training process, Figure \ref{fig:Loss} shows the training and test loss curves on ModelNet40. As can be observed, after around 2000 iteration step, the model starts to converge on the training set. The fast convergency behavior means the model is easy to train. In addition, with the increase of iteration, the line for test loss also drops to converge with the line for training set, which indicates there is no overfitting.
	
	\begin{table*}[ht]
		\caption{Visualization of training process}
		\centering
		\setlength{\tabcolsep}{15.2pt} 
		\begin{tabular}{ccccccc}
			\toprule
			G.T. & Step 0 & Step 100 & Step 500 & Step 2,000 & Step 10,000 & Step 39,000\\
			\midrule
			\includegraphics[width=0.08\textwidth]{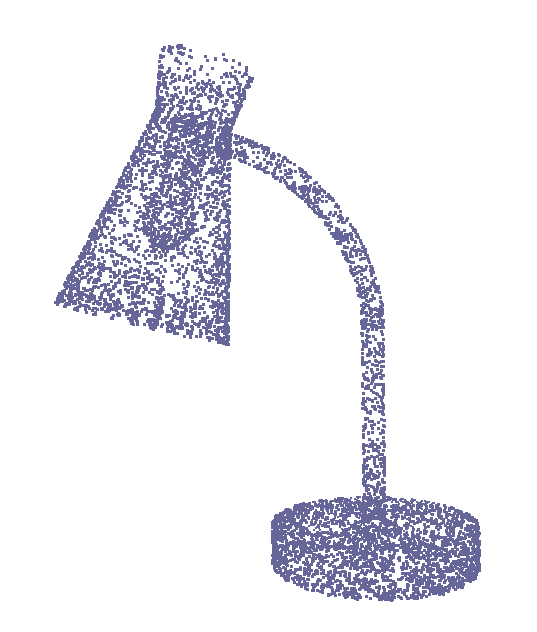} &
			\includegraphics[width=0.08\textwidth]{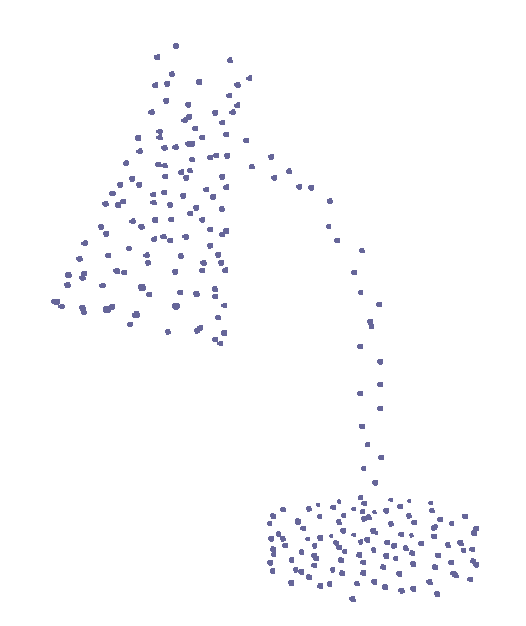} &
			\includegraphics[width=0.08\textwidth]{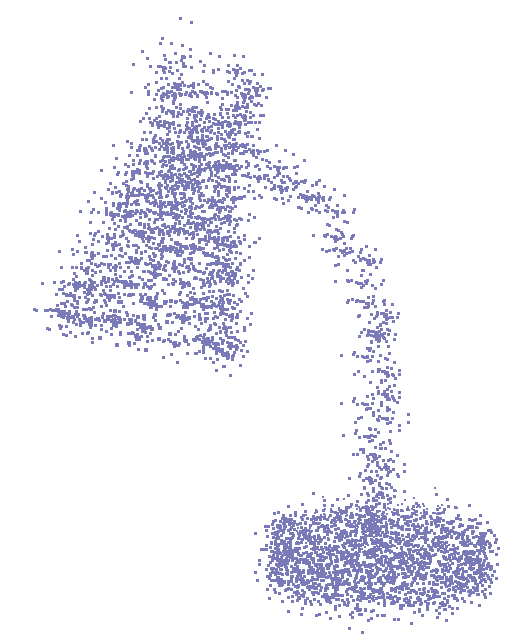} &
			\includegraphics[width=0.08\textwidth]{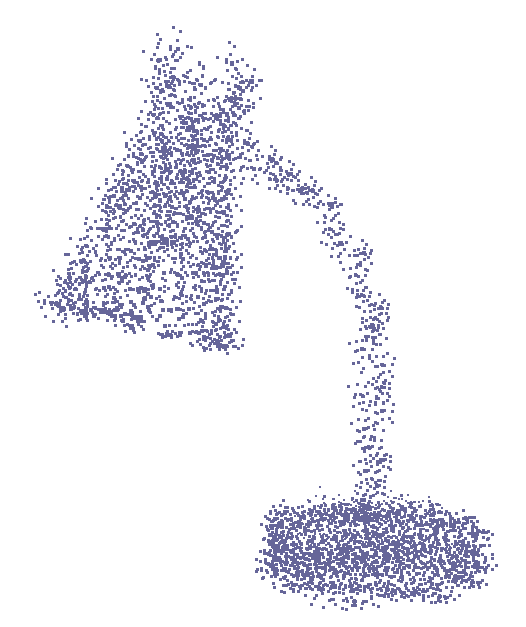} &
			\includegraphics[width=0.08\textwidth]{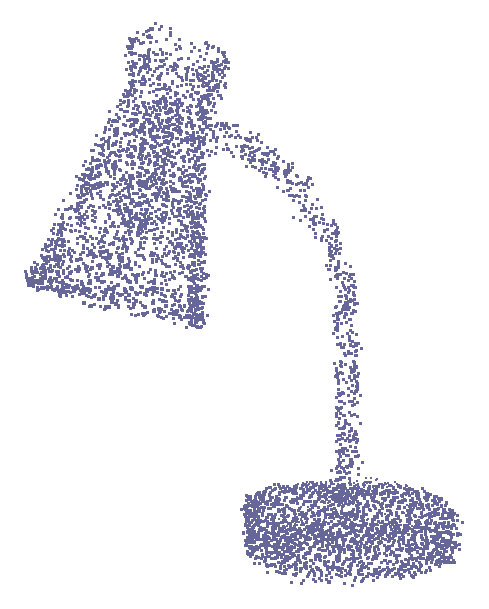} &
			\includegraphics[width=0.08\textwidth]{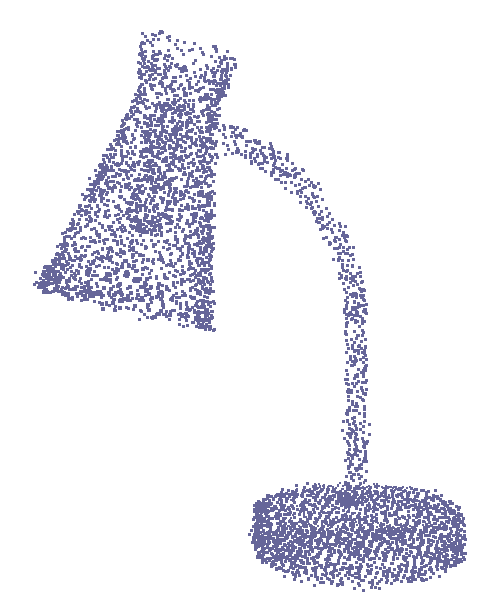} &
			\includegraphics[width=0.08\textwidth]{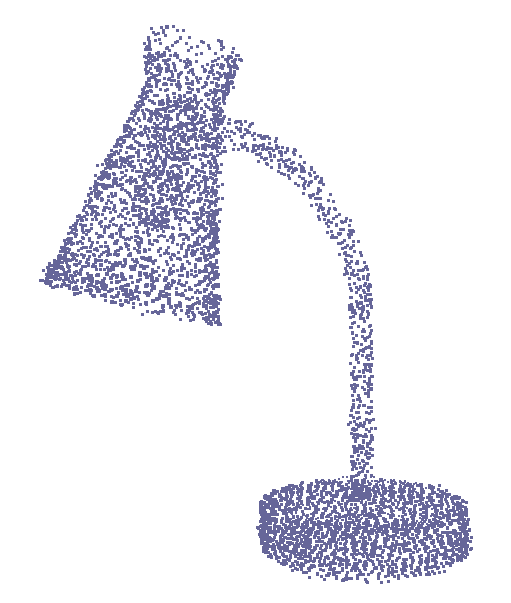} \\
			\includegraphics[width=0.08\textwidth]{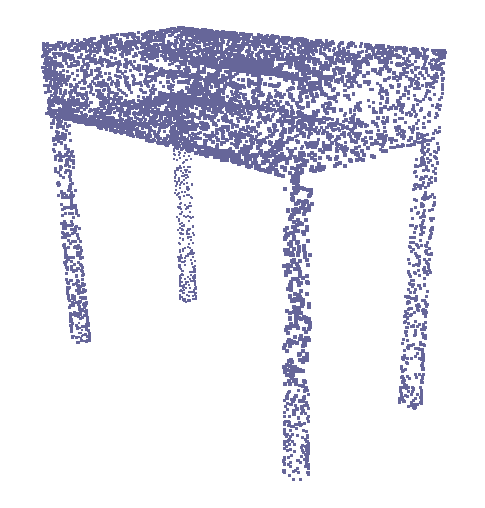} &
			\includegraphics[width=0.08\textwidth]{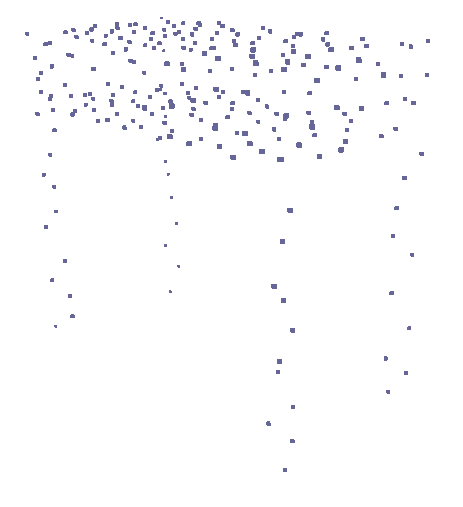} &
			\includegraphics[width=0.08\textwidth]{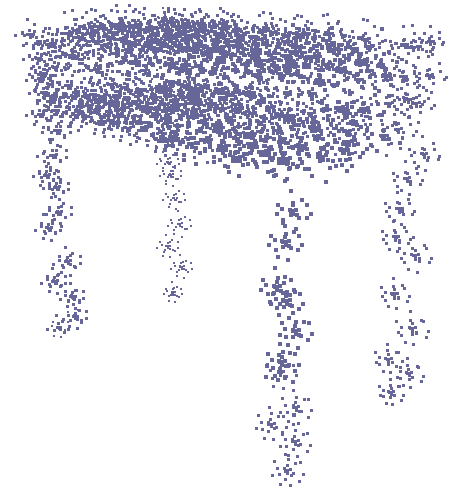} &
			\includegraphics[width=0.08\textwidth]{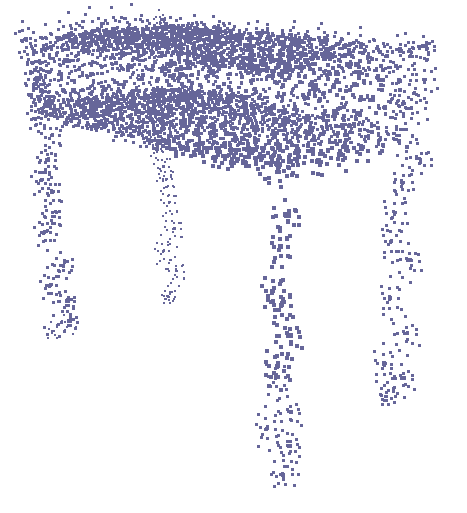} &
			\includegraphics[width=0.08\textwidth]{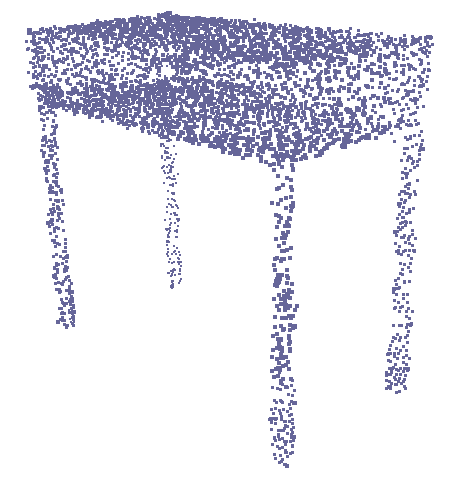} &
			\includegraphics[width=0.08\textwidth]{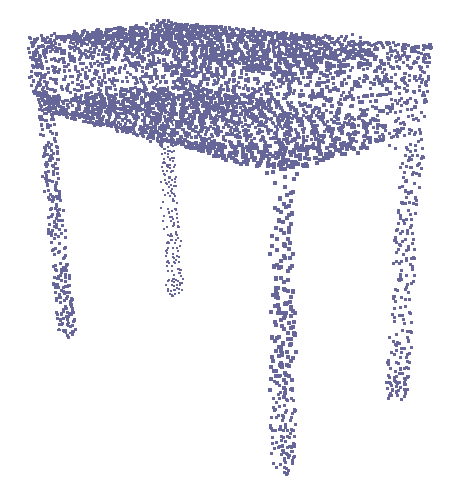} &
			\includegraphics[width=0.08\textwidth]{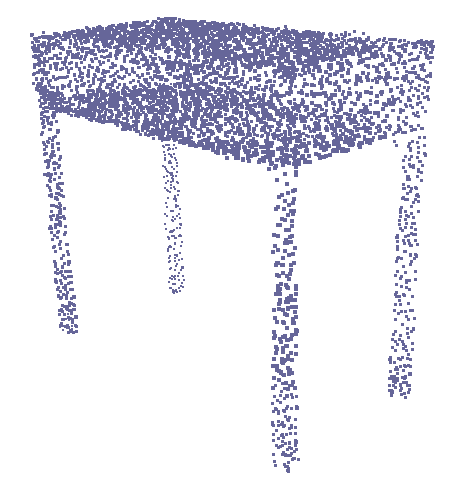} \\
			\bottomrule
		\end{tabular}
		\label{tab:train_process}
	\end{table*}
	
	\subsection{Compression Performance Comparison}
		
	\begin{table}
		\caption{Comparison of Reconstructions of Different Compression Method}
		\centering
		
		\begin{tabular}{cccccc}
			\toprule
			G.T. & Octree & TMC13 & Quach's & Yan's & Ours \\
			
			\midrule
			\includegraphics[width=0.125\linewidth]{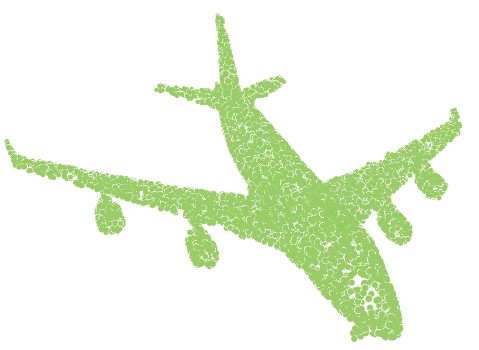} &
			\includegraphics[width=0.125\linewidth]{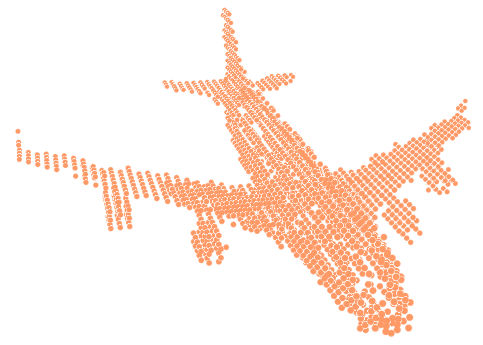} &
			\includegraphics[width=0.125\linewidth]{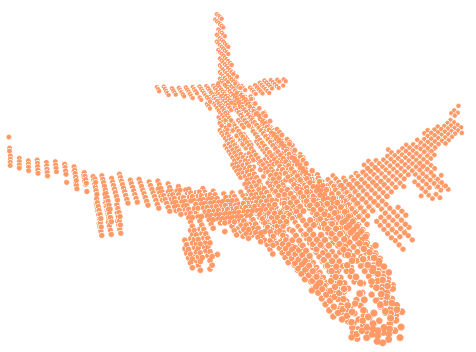} &
			\includegraphics[width=0.125\linewidth]{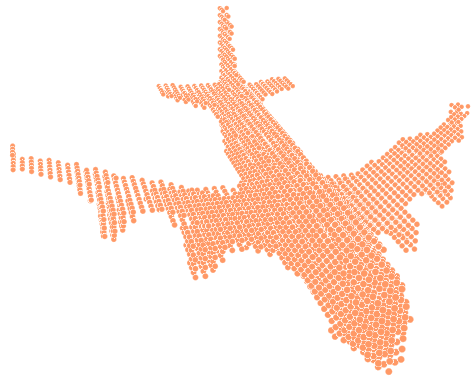} &
			\includegraphics[width=0.125\linewidth]{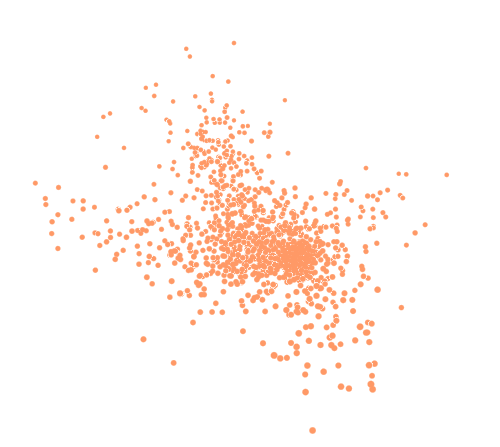} &
			\includegraphics[width=0.125\linewidth]{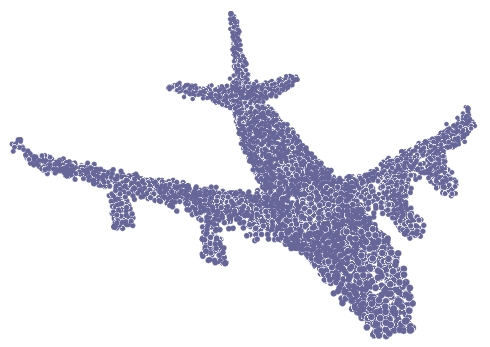} \\
			bpp & 0.843 & 0.630 & 1.358 & 0.168 & 1.114 \\
			PSNR & 13.833 & 9.010 & 8.628 & 0.645 & 18.050 \\

			\includegraphics[width=0.125\linewidth]{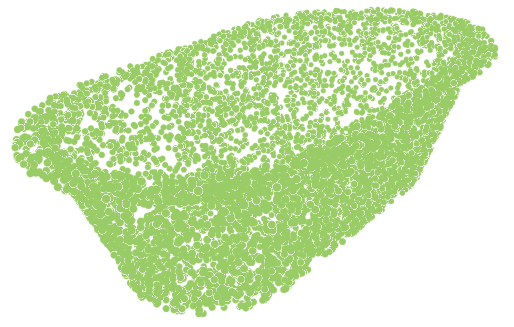} &
			\includegraphics[width=0.125\linewidth]{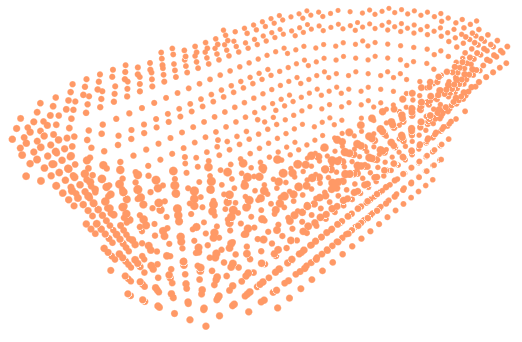} &
			\includegraphics[width=0.125\linewidth]{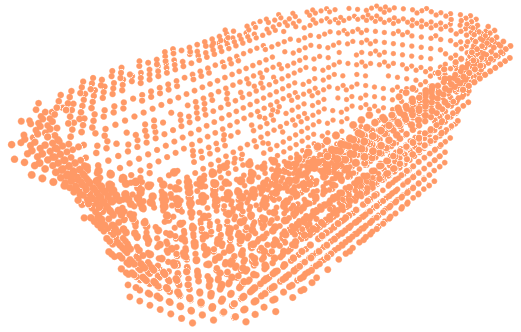} &
			\includegraphics[width=0.125\linewidth]{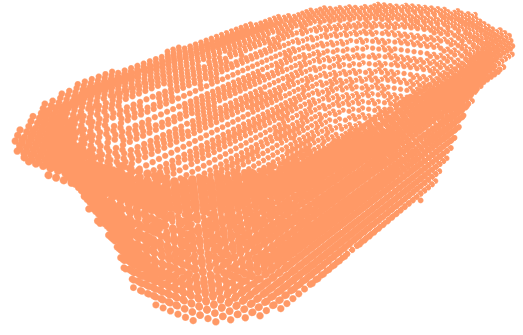} &
			\includegraphics[width=0.125\linewidth]{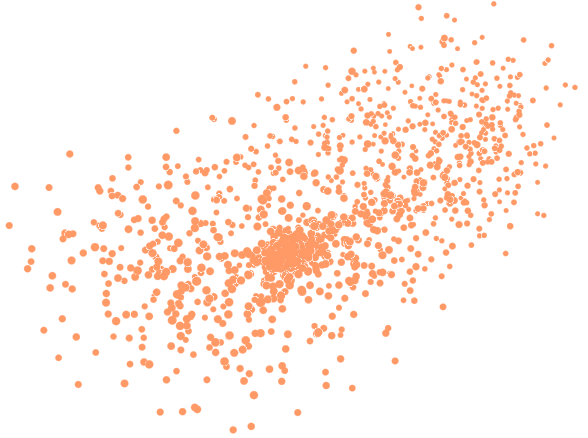} &
			\includegraphics[width=0.125\linewidth]{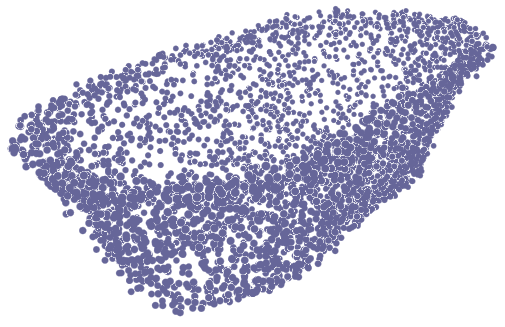} \\
			bpp & 0.489 & 0.619 & 0.509 & 0.203 & 0.480 \\
			PSNR & 7.890 & 8.336 & 13.351 & -1.389 & 13.023 \\

			\includegraphics[width=0.125\linewidth]{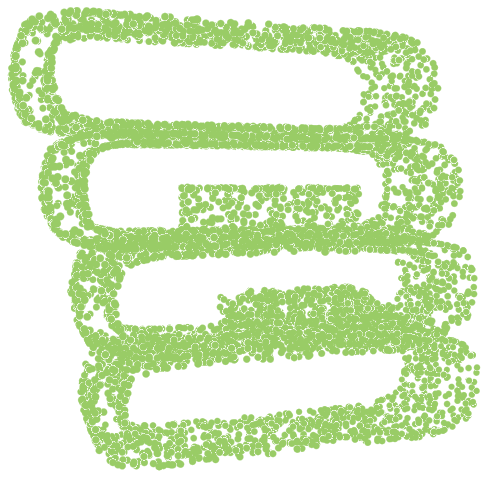} &
			\includegraphics[width=0.125\linewidth]{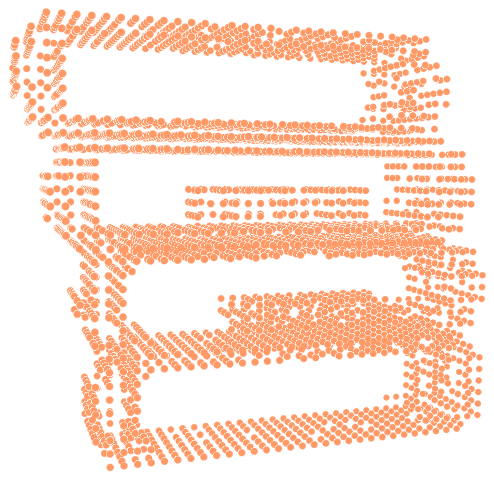} &
			\includegraphics[width=0.125\linewidth]{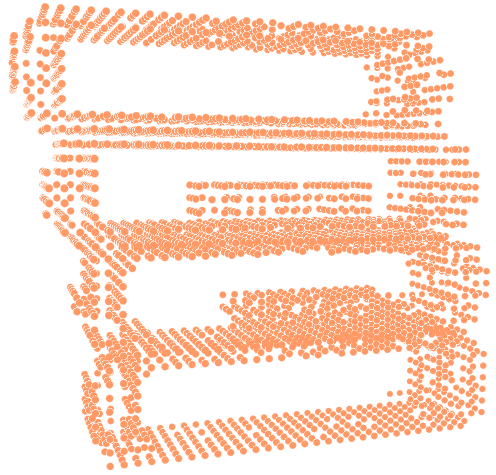} &
			\includegraphics[width=0.125\linewidth]{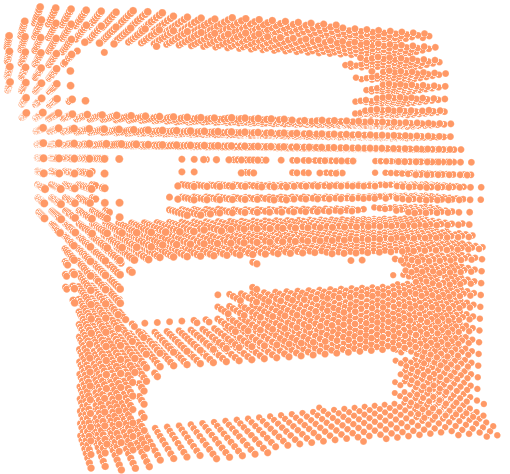} &
			\includegraphics[width=0.125\linewidth]{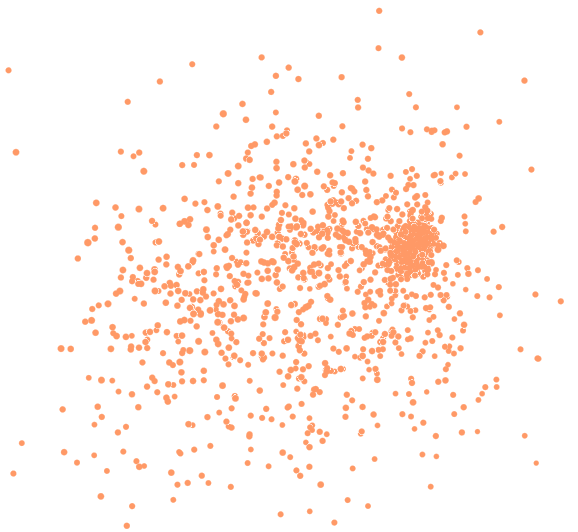}&
			\includegraphics[width=0.125\linewidth]{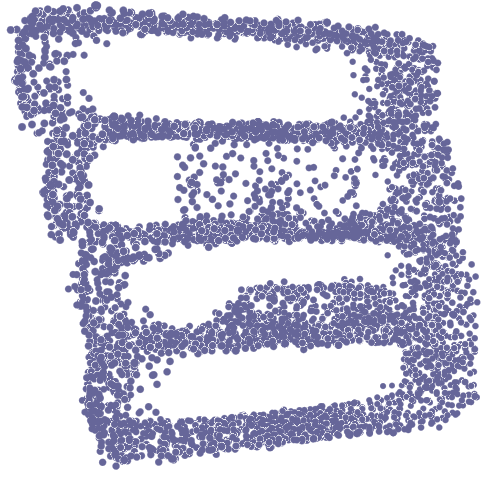}\\
			bpp & 1.100 & 0.906 & 1.140 & 0.172 & 1.168 \\
			PSNR & 15.396 & 11.007 & 8.84 & 4.045 & 15.755 \\

			\includegraphics[width=0.125\linewidth]{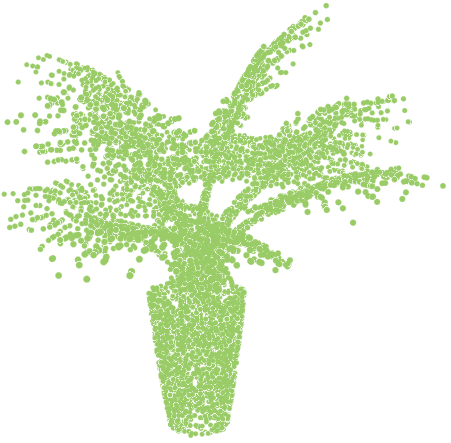} &
			\includegraphics[width=0.125\linewidth]{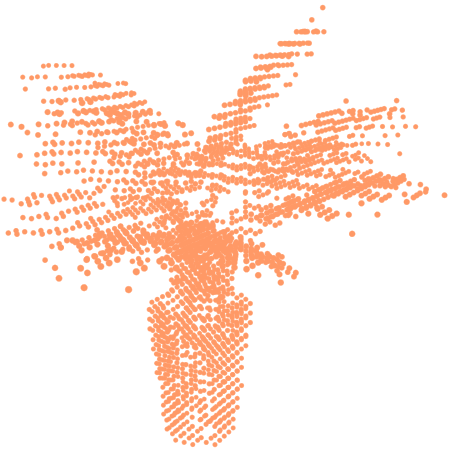} &
			\includegraphics[width=0.125\linewidth]{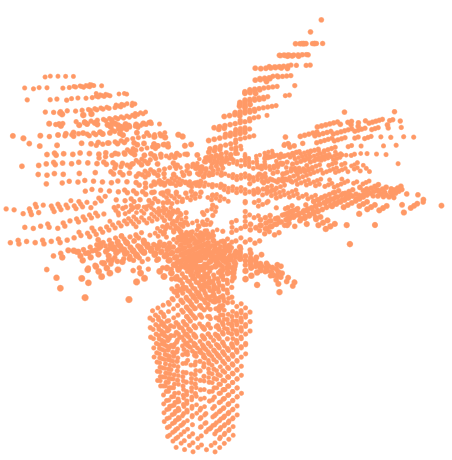} &
			\includegraphics[width=0.125\linewidth]{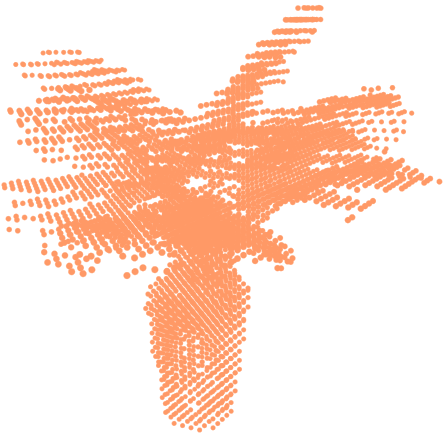} &
			\includegraphics[width=0.125\linewidth]{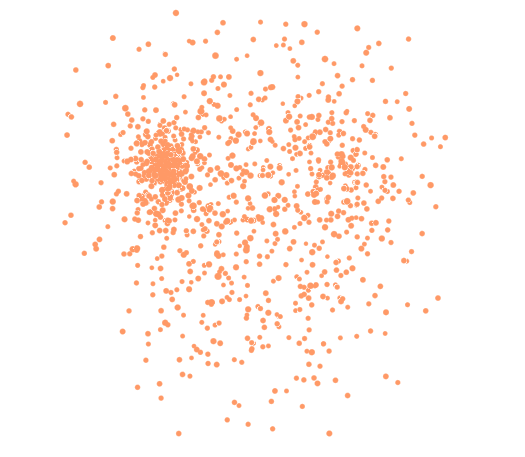} &
			\includegraphics[width=0.125\linewidth]{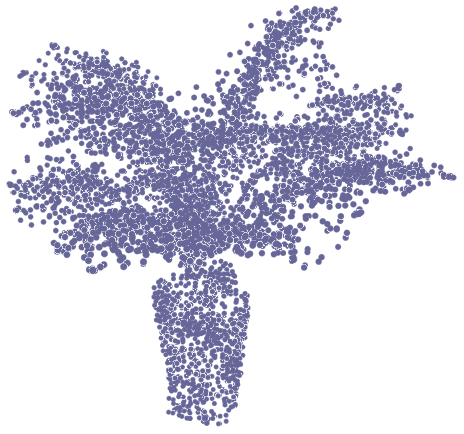}  \\
			bpp & 1.232 & 1.033 & 1.936 & 0.189 & 1.292 \\
			PSNR & 19.494 & 15.065 & 13.15 & 5.300 & 17.661 \\
			
			\bottomrule
		\end{tabular}
		
		\label{tab:comp_reconstruction}
	\end{table}
	
	\begin{figure}[thb]
		\centering
		\subfloat[ModelNet40]
		{\label{fig:c_p_1}\includegraphics[width=0.375\textwidth]{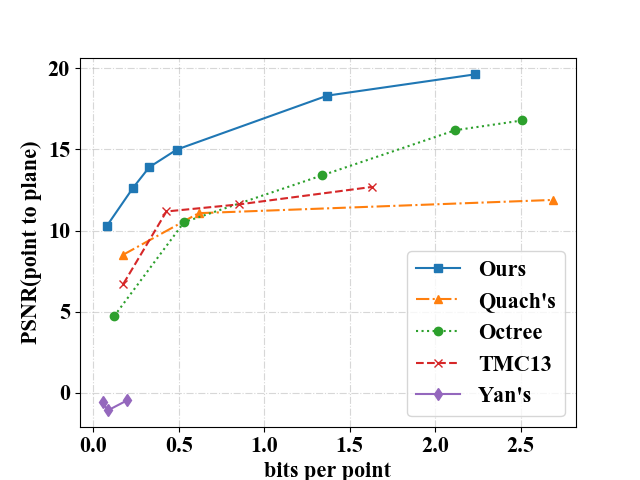}}
		
		\subfloat[ShapeNet]
		{\label{fig:c_p_2}\includegraphics[width=0.375\textwidth]{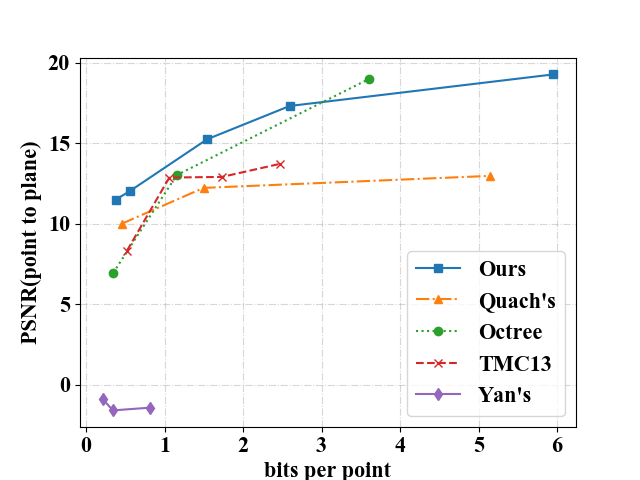}}
		\caption{Compression performance comparison.}
		\label{fig:comp_perforamce}
	\end{figure}
	We compare our method with some representative methods in this area including Octree \cite{Octree}, MPEG recently released static point cloud codec TMC13 \cite{TMC13}, Quach’s voxel-based autoencoder \cite{Quach}, and Yan's deep autoencoder which directly uses PointNet on entire point clouds \cite{DAE}.
	
	To obtain our RD curves, we change patch generation parameters and bottleneck size, i.e., $S$, $K$, and $d$. Since we train our network on the 8192-point resolution point cloud data, we use the same data to train Quach and Yan's network. As in \cite{Quach}, we get Quach’s RD curves with different $\lambda$. We train Yan's network with the same network parameters described in \cite{DAE}, and change its bottleneck size to get RD curves. In TMC13, we use the default octree-based scheme for geometry compression. By setting different quantization parameters for geometry coding, we can obtain various rate distortion points.
	
	Figure \ref{fig:comp_perforamce} shows the RD curves on ModelNet40 8192-point resolution test set and ShapeNet 2048-point resolution test set. We can see that we have significantly better compression performance than Octree, TMC13, Quach's and Yan's on 8192-point resolution point cloud. For point clouds with 2048 points, our method still shows excellent robustness. Due to the naive use of PointNet, Yan's method generally has the lowest RD performance. Note that, the rate distortion curves are obtained by averaging the coding results of all the point cloud examples in the test set.

	Table \ref{tab:comp_reconstruction} shows reconstruction quality comparison of several point clouds in ModelNet40. It can be seen that our method can generate uniformly distributed point clouds with the same resolution as input. Yan's autoencoder completely fails to reconstruct the complex shapes in ModelNet40. It is worth pointing out that, for Yan's method, we directly use the codes provided by the authors in \cite{Deep_AE_github} to train the model and test. The main reason for the poor performance is, since PointNet is originally designed for 3D classification, straightforwardly applying it on entire point cloud feature extraction for compression may result in very unsatisfactory results \footnote{In the original paper \cite{DAE}, more visually plausible reconstruction results are provided. This is because, they train the model class-by-class, i.e., each class point clouds (e.g., chair) has been trained with a model. However, in this comparison, we use all the class  data together in ModelNet40 to train the model, which is a more general way.}. 
	Voxelization inevitably has coordinate offset error. The effect of voxelization on point cloud quality is reflected in the results of Quach's method, in which some details of point cloud have not been properly reconstructed, for example, the leaves of the plant in the last row in Table \ref{tab:comp_reconstruction}. In contrast, without voxelization, we can get each reconstructed point closer to the ground truth.
	
	\subsection{Influence of Patch Count on Compression Performance}
	
\begin{figure}[t!]
		\centering
		\includegraphics[width=0.375\textwidth]{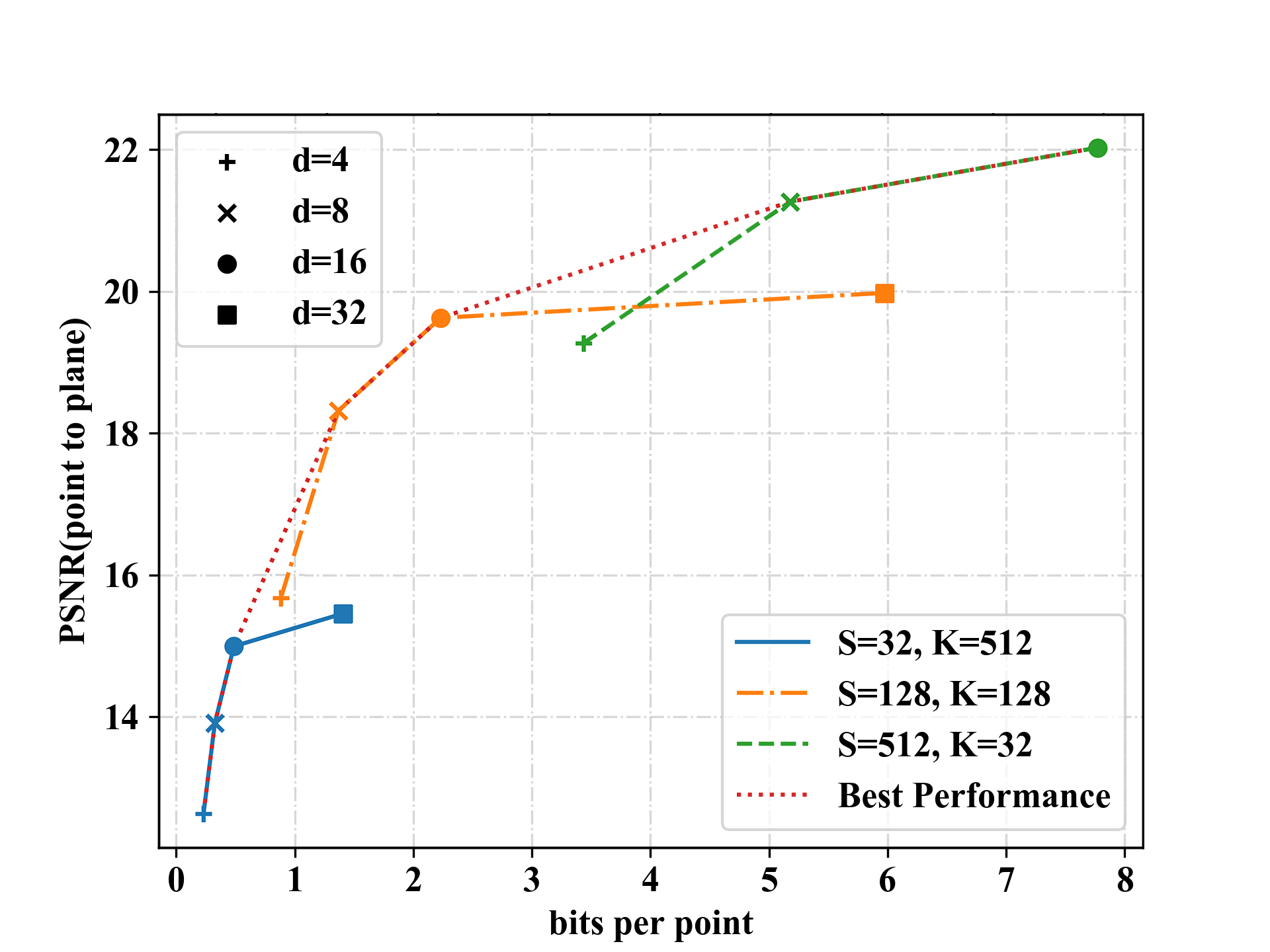}
		\caption{Compression performance curve comparison by using different $S$ and $K$ on ModelNet40.}
		\label{fig:SK_influence}
	\end{figure}

	We tested the effect of different patch dividing parameters on compression performance. We tested a few fixed sets of $S$ and $K$, and then obtained the RD curves by applying different bottleneck sizes. As shown in Figure \ref{fig:SK_influence}, the experimental results show that the compression performance can be constrained because either the bottleneck size is too large or too small. For the 8192-point resolution point cloud, our method can achieve the best performance when $d$ is around 8 to 16. And it also can be concluded from the figure that large-resolution patches are suitable for low bit-rate compression, while small-resolution patches are more suitable for high-quality reconstruction at high bit rates.
	
	\subsection{Application to Point Cloud Upsampling}
	
	\begin{table}[t!]
		\caption{Upsampling Result Illustration.}
		\centering
		\begingroup
		\setlength{\tabcolsep}{8pt} 
		\begin{tabular}{cccc}
			\toprule
			Input (4096 points) &
			\includegraphics[height=0.065\textwidth]{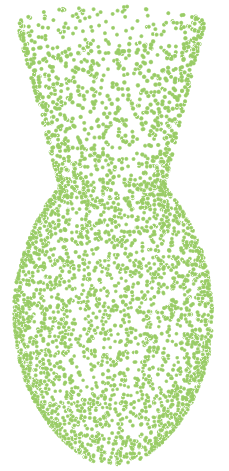} &
			\includegraphics[height=0.065\textwidth]{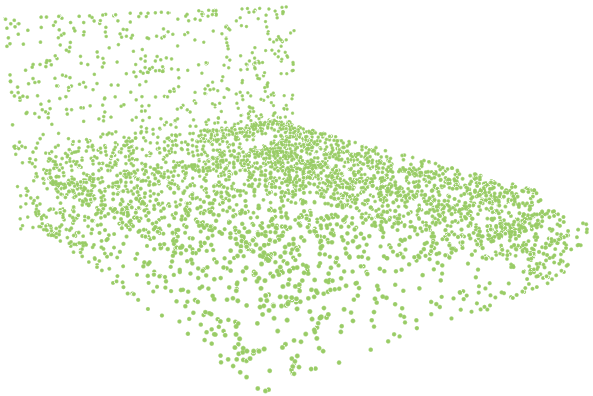} &
			\includegraphics[height=0.065\textwidth]{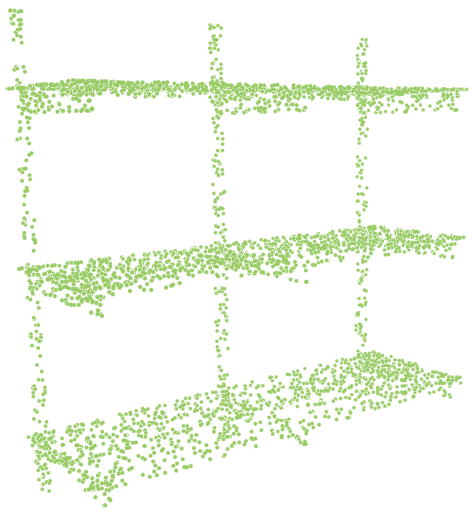} \\
			
			\midrule
			Output (32768 points) &
			\includegraphics[height=0.065\textwidth]{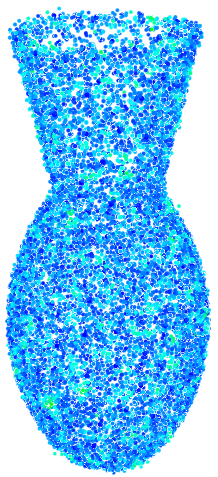} &
			\includegraphics[height=0.065\textwidth]{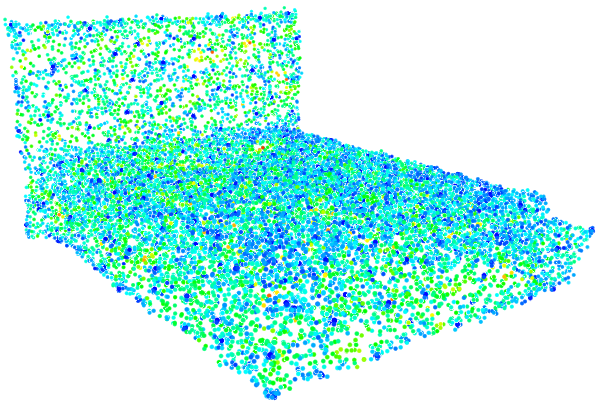} &
			\includegraphics[height=0.065\textwidth]{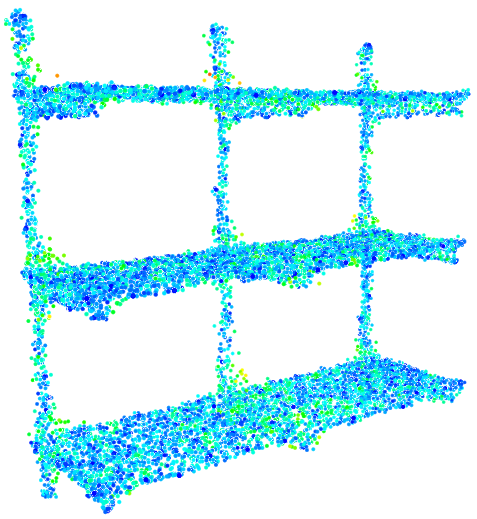} \\
			
			\bottomrule
		\end{tabular}
		\endgroup
		\label{tab:upsampling}
	\end{table}
	
	By exploiting the idea of patch-to-patch reconstruction in our proposed autoencoder, we designed a structure for upsampling, in which we remove the quantization and entropy coding process in the compression architecture in Figure \ref{fig:nn_archi}. And by training the established upsampling networks with Chamfer distance loss, we can improve the resolution of the input point cloud as high as we want. Unlike our compression process, we set $S \times K = \alpha N $ and $ k = M \times K$ to generate the point cloud with a resolution of $M\times \alpha$ times the original point cloud. The results are illustrated in Table \ref{tab:upsampling}. We find that the upsampling model crafted from the proposed autoencoder outputs the upsampled point cloud with good uniformity and surface.
	
	Specifically, we implement our upsampling network using the following parameters:
	
	\begin{math}
	    SA\left ( K, 8, \left[ 32, 64, 128 \right] \right )\rightarrow PN \left( \left[ 256, 512, d \right] \right) \rightarrow FC \left( d, 1024 \right)\rightarrow ReLU \rightarrow FC \left( 1024, 512 \right) \rightarrow ReLU \rightarrow FC \left( 512,M \times K \times 3 \right)
	\end{math}

	\section{Conclusion}
	\label{sec:Conclusion}
	We present a patch-based approach for lossy point cloud geometry compression using deep learning. By using patch division for point cloud, our method changes the global reconstruction problem into a local reconstruction problem, and uses the local reconstruction loss for optimization to approximate the global optimization. This design allows for PointNet to capture internal structure inside point clouds more efficiently, and also brings an effective training data augmentation. Our approach outperforms the existing state-of-the-art methods developed in point cloud compression. The idea of this method can be readily extended to other point cloud reconstruction problems. Source code demonstrating our system is available at \url{https://github.com/I2-Multimedia-Lab/PCC_Patch}.
	


\bibliographystyle{ACM-Reference-Format}
\bibliography{mybib}

\end{document}